\documentclass{article}
\usepackage{float}
\PassOptionsToPackage{numbers, compress}{natbib}
\usepackage[final]{unireps_2023_arxiv}
\usepackage[utf8]{inputenc} 
\usepackage[T1]{fontenc}    
\usepackage{hyperref}       
\usepackage{url}            
\usepackage{booktabs}       
\usepackage{amsfonts}       
\usepackage{nicefrac}       
\usepackage{microtype}      
\usepackage{xcolor}         

\usepackage{graphicx}

\title{ReWaRD: Retinal Waves for \\ Pre-Training Artificial Neural Networks \\ Mimicking Real Prenatal Development}

\author{%
Benjamin Cappell \\
\texttt{benjamin.cappell@fau.de} \\
\And 
Andreas Stoll \\
\texttt{andi.stoll@fau.de} \\
\And
Williams Chukwudi Umah \\
\texttt{chukwudi.umah@fau.de} \\
\And
Bernhard Egger \\
\texttt{bernhard.egger@fau.de} \\
\\
Cognitive Computer Vision Lab, Chair of Visual Computing \\
Friedrich-Alexander-Universität Erlangen-Nürnberg, Germany
}

\begin{document}
\maketitle
\begin{abstract}
  Computational models trained on a large amount of natural images are the state-of-the-art to study human vision -- usually adult vision. Computational models of infant vision and its further development are gaining more and more attention in the community. In this work we aim at the very beginning of our visual experience -- pre- and post-natal retinal waves which suggest to be a pre-training mechanism for the primate visual system at a very early stage of development. We see this approach as an instance of biologically plausible data driven inductive bias through pre-training. We built a computational model that mimics this development mechanism by pre-training different artificial convolutional neural networks with simulated retinal wave images. The resulting features of this biologically plausible pre-training closely match the V1 features of the primate visual system. 
  We show that the performance gain by pre-training with retinal waves is similar to a state-of-the art pre-training pipeline. Our framework contains the retinal wave generator, as well as a training strategy, which can be a first step in a curriculum learning based training diet for various models of development. 
  We release code, data and trained networks to build the basis for future work on visual development and based on a curriculum learning approach including prenatal development to support studies of innate vs. learned properties of the primate visual system. An additional benefit of our pre-trained networks for neuroscience or computer vision applications is the absence of biases inherited from datasets like ImageNet.
\end{abstract}

\begin{figure}[H]
  \centering
  \includegraphics[width=\textwidth]{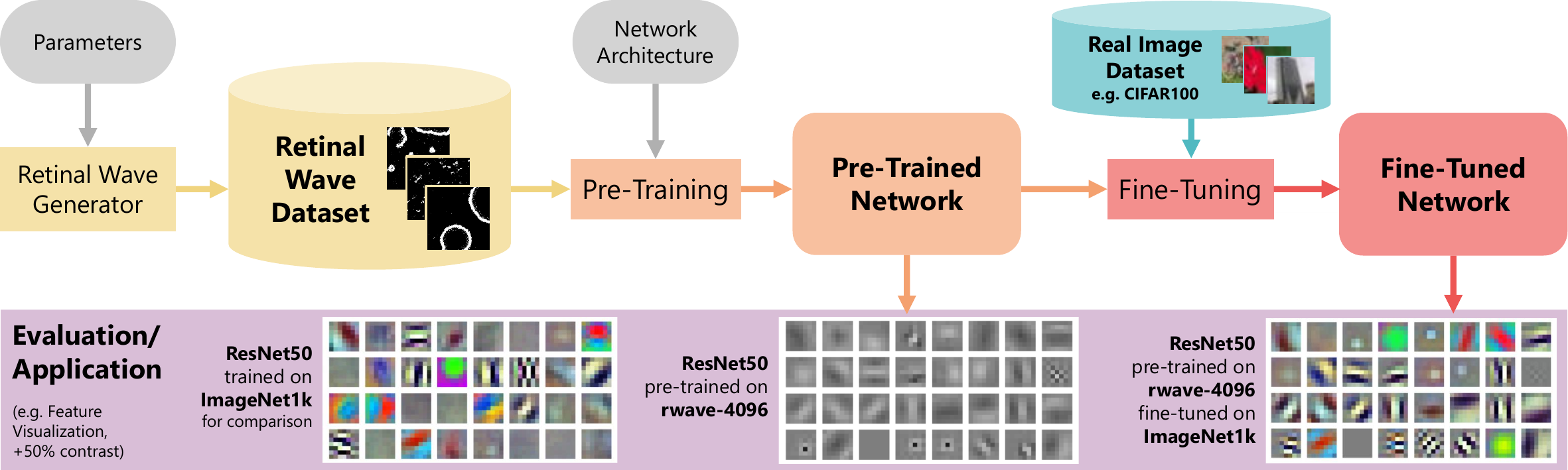}
  \caption{Overview of our framework including retinal wave generation, pre-training and fine-tuning. \\ Feature visualizations show first convolutional layer.}
\end{figure}

\section{Introduction}
Retinal waves are a phenomenon of randomly initiated wave-like patterns, which travel over the developing retina, for mammals first reported by Meister et\ al.\ in 1991 \cite{Meister1991}. They are difficult to observe, e.g.\ using calcium imaging and multielectrode array recordings \cite{Firth2005}, as they occur prenatally before eye opening and cone/rod maturation. Retinal waves play a role in the development of neural pathways (in and between retina, LGN and V1). One example is the development of long range horizontal connections, affecting the orientation map in the visual system based on the directions of retinal waves \cite{Kim6584}.

At different stages of the development, retinal waves are initiated in various biological methods. We focus on earlier, cholinergic amacrine cell mediated waves \cite{Firth2005}. Reported retinal waves differ in shape, size and propagation speed. This variety is what could be a strong basis to interpret retinal waves as input for prenatal development of the visual system.

In pre-natal mice, directional retinal waves have been shown to simulate future optical flow, therefore priming visual motion detection even before the onset of vision. This suggests the development of higher-order visual processing regions prior to eye opening \cite{ge2021retinal}. 

We assume especially the early layers of visual processing to profit from pre-training through retinal waves, as those are known to capture basic features similar to Gabor features. Later regions like V4 or IT might get an initial signal for wiring and feature learning, but those regions are more object-centric and therefore likely profit less from this first visual experience.
At eye opening, the wiring of the visual system is initialized and can be refined by utilizing visual impressions of the real world. 

Similar to a biological pre-training of the visual system with retinal waves, pre-training of Artificial Neural Networks (ANNs) (even with non-natural images) is used to generalize faster and to boost accuracy \cite{KataokaACCV2020, KataokaIJCV2022}. 

In our framework, we perform training in a fully supervised setting. This is possible by assigning an individual class label to each wave. This comes with the heuristic, that a retinal wave is processed in a somewhat holistic way and therefore such a supervision signal (e.g. current stage of the development process) could be also available during training of a biological system. Whilst unsupervised training would be an option with our synthetic retinal wave data, we focus on supervised learning, as this has been shown to match brain activity better on various benchmarks.

Related work has shown that Slow Feature Analysis (SFA) units derived from simulated retinal waves share a number of properties with cortical complex-cells \cite{Dhne2014} and that pre-training a classifier ANN with real and simulated retinal waves using a Hebbian learning rule improves separability of NN-internal representations and classification accuracy \cite{ligeralde2022geometry}.

This leads us to the following research questions: 
\begin{itemize}
    \item What features would arise in ANNs if we mimic development of the visual system through retinal waves by pre-training ANNs with simulated retinal wave images?
    \item How similar will these ANNs be to the human brain internally?
    \item How well will these networks perform when fine-tuned for image classification tasks?
\end{itemize} 

In summary the core contributions of our work are as follows:
\begin{itemize}
\item We adapt an existing retinal wave simulator for the purpose of image generation for pre-training ANNs and create two different retinal wave datasets.
\item We pre-train and fine-tune ANNs in various different settings and evaluate them qualitatively and quantitatively in terms of how well they match biological visual systems and how they compare to another pre-training approach with generated images of fractals \cite{KataokaACCV2020} \cite{KataokaIJCV2022}.
\item We release all datasets, code and pre-trained networks to foster the development of variants of our experiments. Therefore, custom retinal wave based curriculum learning strategies may be built upon our pre-trained networks.
\end{itemize}
The paper is structured as follows: we first describe the retinal wave generator and the particular datasets we generated. Second, we provide details about our pre-training and fine-tuning scheme. Finally we present different qualitative and quantitative evaluations of our method by showing the learned features, measuring accuracy on an object recognition task and comparing how similar the activations in our models are to activations measured in biological vision systems.

\begin{figure}[H]
    \centering
    \includegraphics[width=0.618\linewidth]{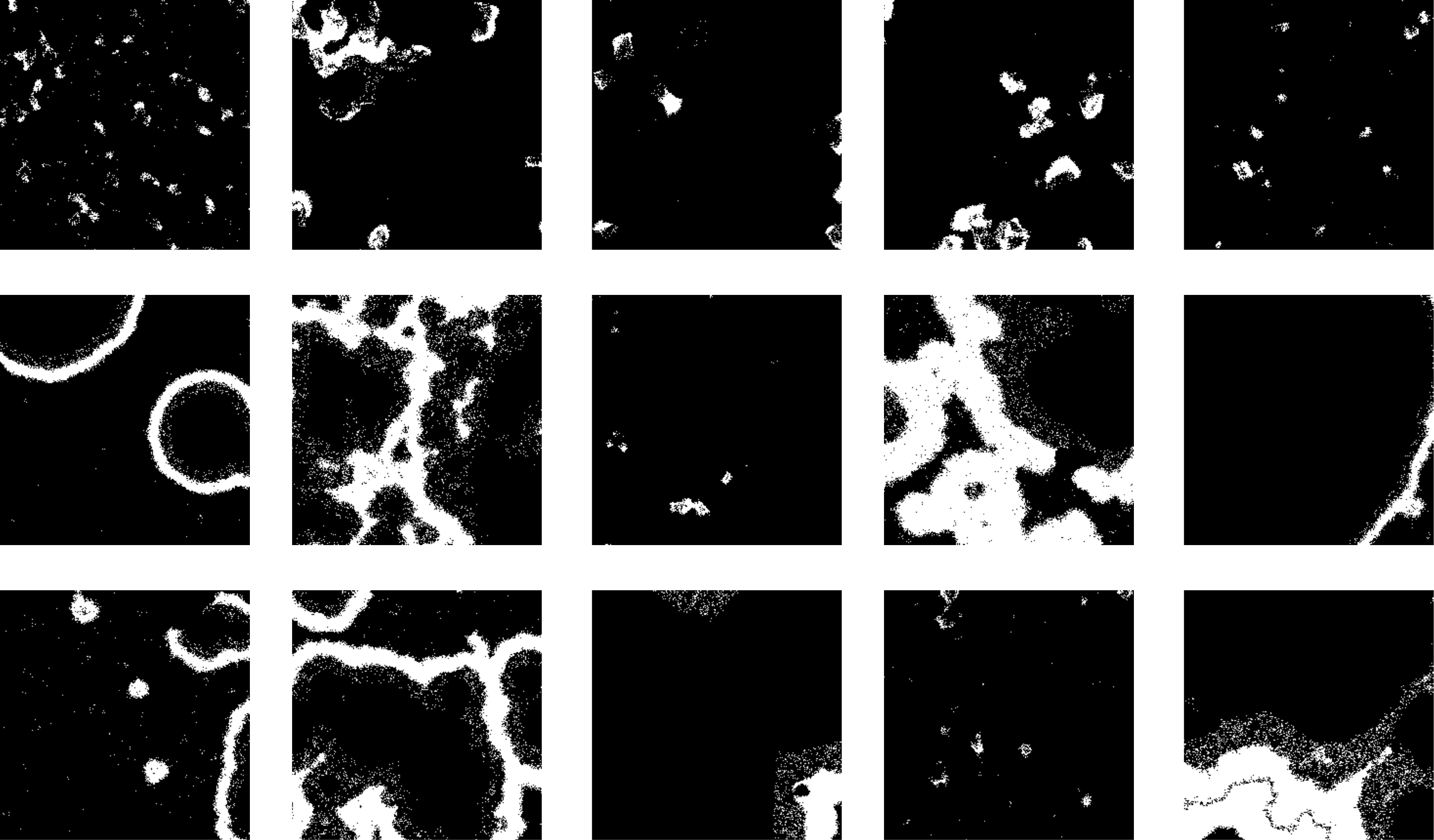}
    \caption{Simulated retinal wave images (random examples from rwave-4096)}
    \label{fig:fig2a}
\end{figure}

\section{Retinal Wave Datasets}

To obtain a large amount of retinal wave data required to pre-train ANNs, we modified a biologically accurate retinal wave simulator \cite{Godfrey2007}.
It simulates and renders retinal waves by outputting activations of dendritically interconnected amacrine cells. These cells are organized in a hexagonal grid and fill a circle with specifiable radius.

Sample parameter sets for the amacrine cells are available to closely reproduce observed retinal waves of different species during different stages of development \cite{Godfrey2007}. 
Every time frame of a retinal wave, whose propagation, shape and size is influenced by multiple adjustable parameters of the amacrine cells, can be simulated. The parameters we control affect the wave shape, speed, duration and size. For further details about these parameters we refer the interested reader to \cite{Godfrey2007}.

Our modifications to the retinal wave simulator include: Generating datasets by choosing base parameters of amacrine cells and specifying parameter spread, which leads to multiple parameter combinations; Retina-to-image projection of retinal waves; Modifying the simulator to support larger retinas; Storing generated retinal waves as .png image files. In Figure~\ref{fig:fig2a}, exemplary images from generated waves are visualized.

\begin{figure}
    \centering
    \includegraphics[width=0.618\linewidth]{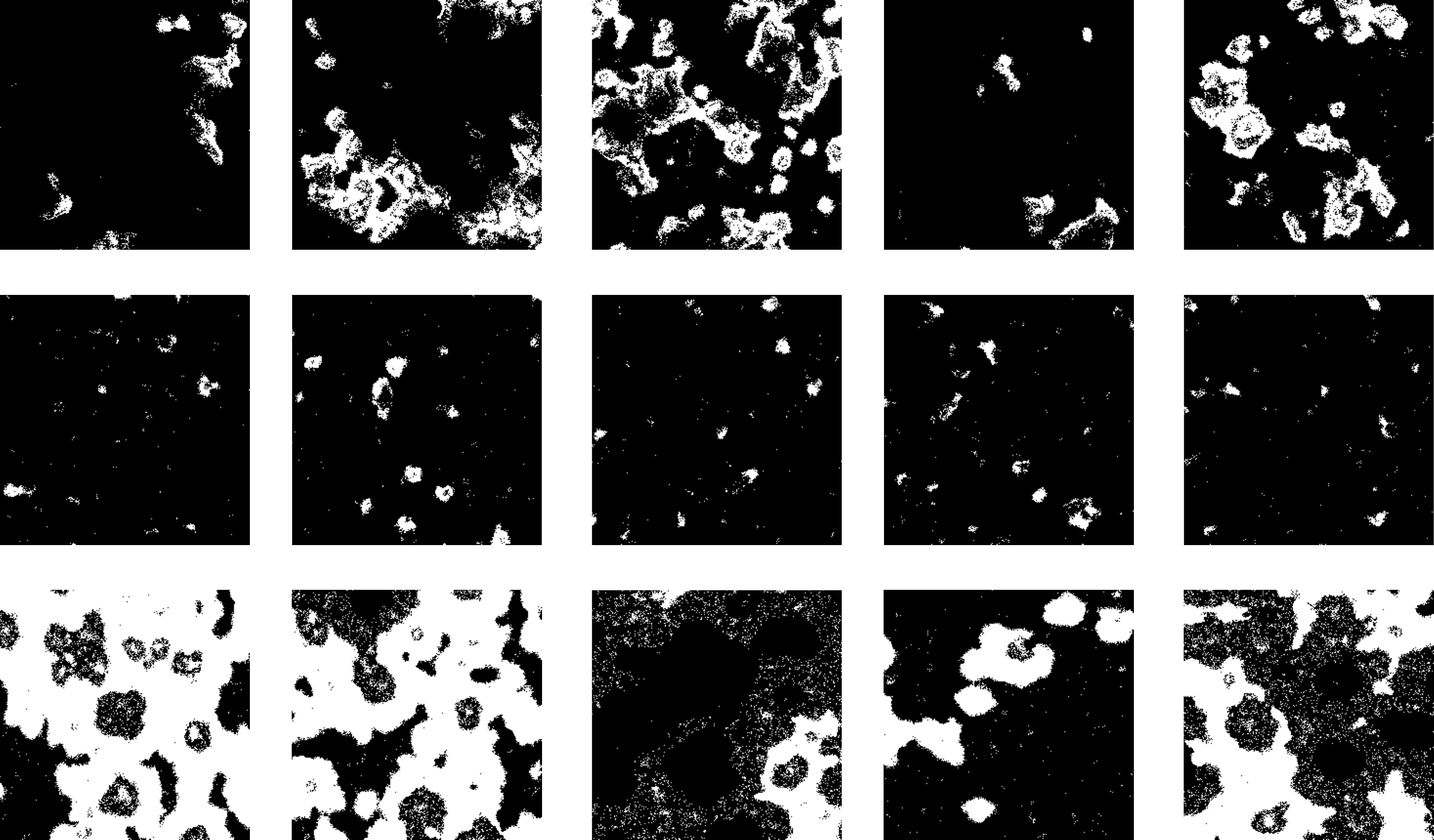}
    \caption{Simulated images: random examples of three classes (rows). We generate a set of retinal waves based on different parameters and each wave is given a separate class label. Different time frames of the same retinal wave are assigned the same class label. This enables us to train ANNs in a fully supervised fashion.}
    \label{fig:fig2b}
\end{figure}

Different parameter combinations for retinal waves act as different classes to train a network in a fully supervised setting. As the parameters are continuous, an unlimited number of classes can be created. A specific parameter constellation corresponds to one class for the supervised learning task. By having multiple frames of a retinal wave, each class contains multiple image frames from the same wave acquired at different time steps. Instances of three different example classes are shown in Figure~\ref{fig:fig2b}.

With this approach, an arbitrary number of different image instances can be generated for each class. This enables us to generate training data for a fully supervised learning task, comparable to the one proposed by FractalDB \cite{KataokaACCV2020, KataokaIJCV2022} and comparable to the standard ImageNet based training which leads to high similarity to biological processing.

Not every temporal frame of a retinal wave is used for the image dataset: we skip retinal wave images during dataset creation if they are temporally too close to one another (spacing, only every n-th image of the wave is considered for a dataset) or if they don't contain a specific amount of pixels (threshold). This choice helps to only include images, where retinal waves are clearly visible, keep the size of the training dataset reasonable and have more variance in the dataset.

We used random mirroring and random continuous 360° rotation as data augmentation strategies. If for a particular class the specified amount of images could not be generated due to too high threshold or temporal spacing, threshold and spacing are dynamically adjusted for the affected class, until the desired amount of images is generated for the class.

\begin{figure}[H]
    \centering
    \includegraphics[width=0.46\linewidth]{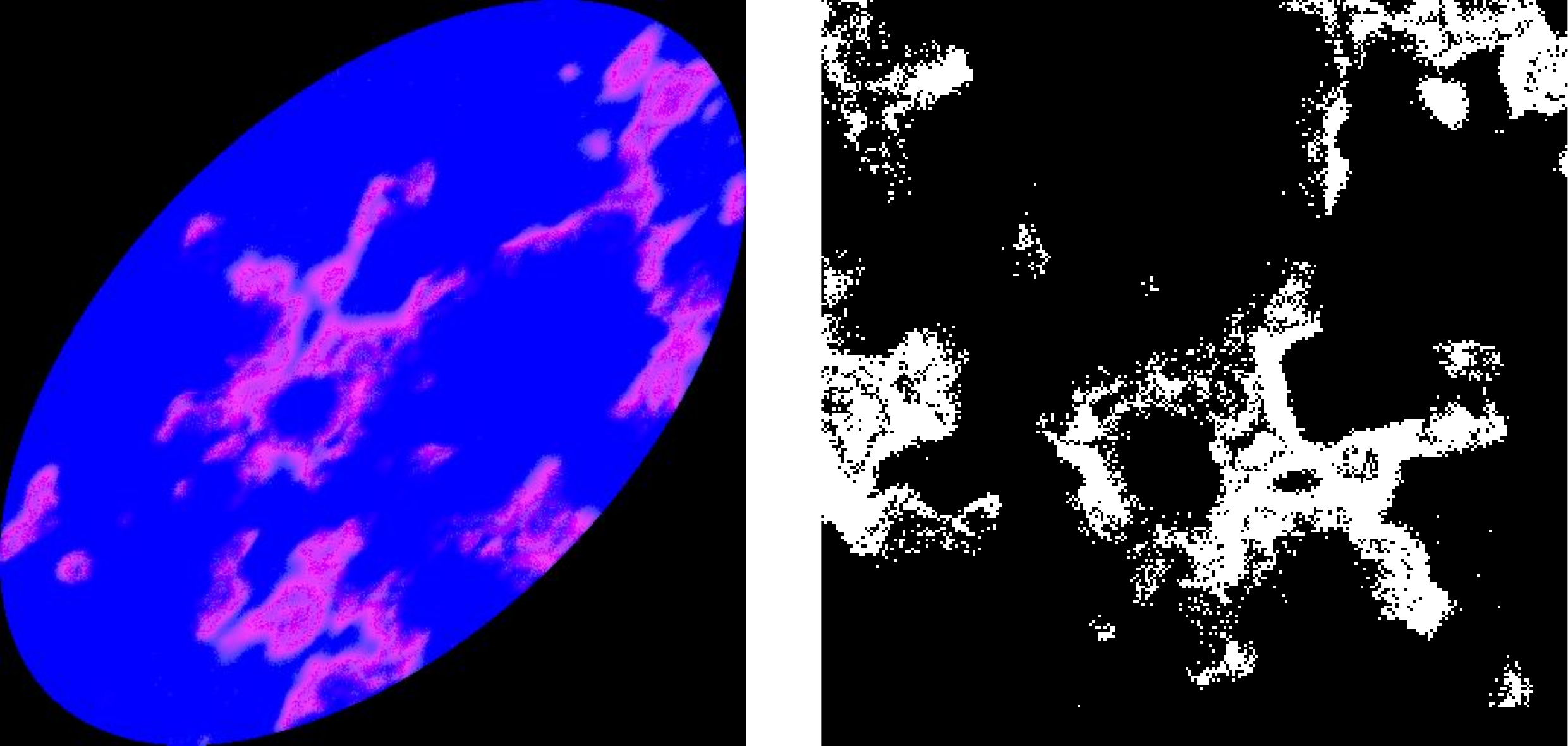}
    \caption{Raw retinal wave image (left); Cropped, shifted and augmented retinal wave image (right). }
    \label{fig:fig2}
\end{figure}

The adapted retinal wave generator can output two types of image data. Firstly, cropped, square images. The corners of the cropped images line up with the retina circle outline such that all pixels of the image represent amacrine cells of the simulated retina. The image size is specifiable, independent of the amount of simulated amacrine cells (default: 256x256 pixels). The resulting images are binary .png files, 1 = cells active, 0 = cells inactive. An example image is shown in Figure~\ref{fig:fig2} (right).

Secondly, images containing raw, unresampled retinal data can be outputted as 3x8 bit RGB .png images (red channel: simulated calcium imaging response of the retina (8 bit), green channel: raw amacrine cell activity (2 bit), blue channel: retina boundary (1 bit)). An example is shown in Figure~\ref{fig:fig2} (left). These raw images could be a basis to adapt and reuse the retinal wave dataset for recurrent time series learning, as there are no temporal jumps (i.e. every generated frame is stored) and the data is not randomly augmented.

Additionally, for each class a textfile is generated, containing the exact parameter values used for generating retinal waves of that class. Thus, the dataset could also be used for regression tasks.

\pagebreak
\begin{table}
  \caption{Created retinal wave datasets. For each of the parameters (affecting wave shape, speed, duration, spacing and size), four different values were used. The class count results by generating classes for all of the possible parameter value combinations (e.g.\ $4^5=1024$).}
  \label{tab:datasets}
  \centering
  \begin{tabular}{lllll}
    \toprule
    \it dataset     & classes     & images/class & altered params & dimensions \\
    \midrule
    \it rwave-1024 & 1024  & 1000  & 5 & 256x256 pixels  \\
    \it rwave-4096 & 4096  & 2000  & 6 & 256x256 pixels  \\
    \bottomrule
  \end{tabular}
\end{table}

We created two retinal wave datasets that are made publicly available  \cite{cappell2023rwave-1024, cappell2023rwave-4096}. Further details are provided in Table \ref{tab:datasets}. Random images of the rwave-4096 dataset are shown in Figure \ref{fig:fig2a}.

The parameter ranges were set according to example parameter combinations provided by \cite{Godfrey2007} for retinal waves occurring in ferrets, rabbits, mice, chickens and turtles.  

Both datasets are split into training set (80\%), validation set (10\%) and test set (10\%). The split is done balanced - for one set, all classes contain the same number of instances.

\section{Pre-Training ANNs}
By mimicking mammalian development of the visual cortex during ANN training, we would expect similar features to arise as used by the human visual system. As the first visual cues in this development are retinal waves,  we try to mimic the mammalian visual development in ANNs by providing the first visual input to ANNs as retinal waves. We achieve this in the pre-training phase, by training a classifier network on one of the retinal wave datasets. 

We base our pre-training on an existing approach of pre-training with synthetically generated fractals and thus, reuse their codebase \cite{KataokaACCV2020, KataokaIJCV2022}. This is possible, as their training approach is very similar to ours: firstly, train an image classifier neural network on a pre-training dataset (in the case of FractalDB: fractal images, in our case: retinal wave images), later: fine-tune the pre-trained network on the final task-specific image dataset.

In order to start the artificial development process, an ANN architecture is selected and used to create a new model, typically a convolutional neural network. We focused experiments on the ResNet50 Architecture \cite{he2016deep}, however, we also release pre-trained ResNet34 \cite{he2016deep} and AlexNet \cite{krizhevsky2017imagenet} 
networks since these are heavily used in the neuroscience and computational cognitive science community. The training pipeline (reused from \cite{KataokaACCV2020, KataokaIJCV2022}) additionally supports VGG16, VGG19 \cite{simonyan2014very}; ResNet18, ResNet101, ResNet152, ResNet200 \cite{he2016deep}; ResNeXt101 \cite{xie2017aggregated} and Densenet161 \cite{huang2017densely}.

The training set of the selected retinal wave dataset is used to optimize the model's trainable parameters (task: classify one retinal wave image). The validation set can be used to monitor performance and to tune hyperparameters (e.g. learning rate). After 90 epochs, the pre-trained model is saved for fine-tuning.

\section{Fine-Tuning of Pre-Trained Networks}
As soon as one opens their eyes for the first time, different visual cues become apparent to the visual system: The real world! We mimic this by fine-tuning our pre-trained ANN. No weights are frozen/fixed – as in the real world, the visual system (resp. the ANN) heavily adapts after eye opening.  From our perspective this is an instance of a data-driven inductive bias through pre-training.
The last layer (classification) of the ANN is swapped out for a layer with the output size equal to that required for the fine-tuning dataset, to make classification possible.
This pre-trained model is then used in an instance of transfer learning, as the model with all its weights is re-used for fine-tuning (apart from the last layer).
Task-specific and labeled data is now used for object classification like ImageNet \cite{ILSVRC15} or CIFAR \cite{krizhevsky2009learning}, which are also the image datasets used to train current state-of-the-art models for visual perception.
We again reuse the FractalDB codebase for this fine-tuning step. \cite{KataokaACCV2020, KataokaIJCV2022}

\section{Evaluation}
We present different qualitative and quantitative evaluations of our method in the following. All evaluations were carried out using the trained ResNet50 networks if not mentioned otherwise.
\begin{figure}[H]
    \centering
    \includegraphics[width=\columnwidth]{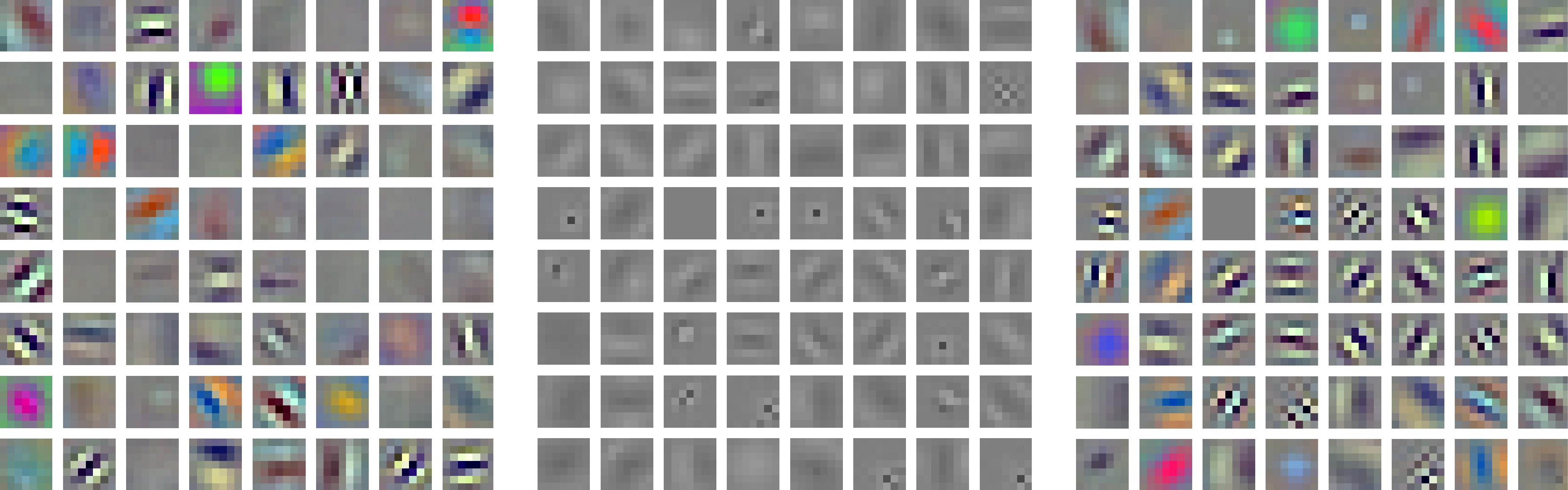}
    \caption{Learned first layer filters of ResNet50 trained from scratch ImageNet1k (left), pre-trained on rwave-4096 (middle) and pre-trained on rwave-4096 + fine-tuned on ImageNet1k (right). For training, the default configuration of the FractalDB pipeline was used. }
    \label{fig:fig3}
\end{figure}

\textbf{Filters:} The learned filters of early layers of pre-trained ResNet50 ANNs closely match gabor-filters, making them similar to the real V1 filters in the human visual system. Similar filters are observed on trained AlexNet networks.
The filters are visualized by showing the weight values. An example is provided in Figure~\ref{fig:fig3}. All visualizations show filters of the first layer of a ResNet50 obtained by training with the default parameters of the FractalDB training pipeline.

The filters of both the pre-trained and fine-tuned networks are similar, as seen in Figure~\ref{fig:fig3}. Most of the gabor-like features are preserved, and increase in contrast or additionally implement color cues.
In comparison to filters of a network only trained on ImageNet1k, there are more gabor-like features on the first layer of a network that was pre-trained with retinal waves and then fine-tuned on ImageNet1k.

\textbf{Accuracy:}
For fine-tuning on CIFAR100, We observe higher accuracies after pre-training on retinal waves and consequently fine-tuning (see Table \ref{tab:acc}), compared with training from scratch. We observe accuracies similar to pre-training with FractalDB1k. Also, we observe that pre-training on ImageNet1k yields the highest accuracies (but at the trade-off of a bias being introduced, as classes and images in ImageNet are hand-picked)
These observations closely matches the results of prior work which was pre-trained on FractalDB \cite{KataokaACCV2020, KataokaIJCV2022}, however, our training diet is biologically plausible.
\begin{table}[H]
    \caption{Accuracies and losses for tested ResNet50 training strategies. For ImageNet1k, the validation set has been used for testing. \textbf{Best} for the last dataset in the training strategy is marked in bold. For these training strategies, the default configuration of the FractalDB training pipeline was used (initial learning rate of 0.01).}
    \label{tab:acc}
    \centering
    \begin{tabular}{rll}
        \toprule
        \it training datasets & Test accuracy $\uparrow$ & Test Loss $\downarrow$ \\
        \midrule
        \it CIFAR100 & 60.29\% & 1.727 \\
        \it rwave-1024 $\to$ \it CIFAR100 & 69.96\% & 1.419 \\
        \it rwave-4096 $\to$ \it CIFAR100 & 72.30\% & 1.330 \\
        \it FractalDB1k $\to$ \it CIFAR100 & 71.22\% & 1.286 \\
        \it ImageNet1k $\to$ \it CIFAR100 & \textbf{79.46\%} & \textbf{0.929} \\
        \midrule
        \it ImageNet1k & 68.23\% &  1.388\\
        \it rwave-4096 $\to$ \it ImageNet1k & \textbf{70.97\%} & \textbf{1.295} \\
        \bottomrule
    \end{tabular}
\end{table}

For fine-tuning on ImageNet1k, we observe the same for the default configuration of the FractalDB training pipeline. Yet, with an initial learning rate of 0.1 during the final training step, training from scratch results in the highest accuracy (see Table \ref{tab:acc2}. This observation matches the results of FractalDB \cite{KataokaACCV2020, KataokaIJCV2022}.

\begin{figure}[H]
    \centering
    \includegraphics[width=\columnwidth]{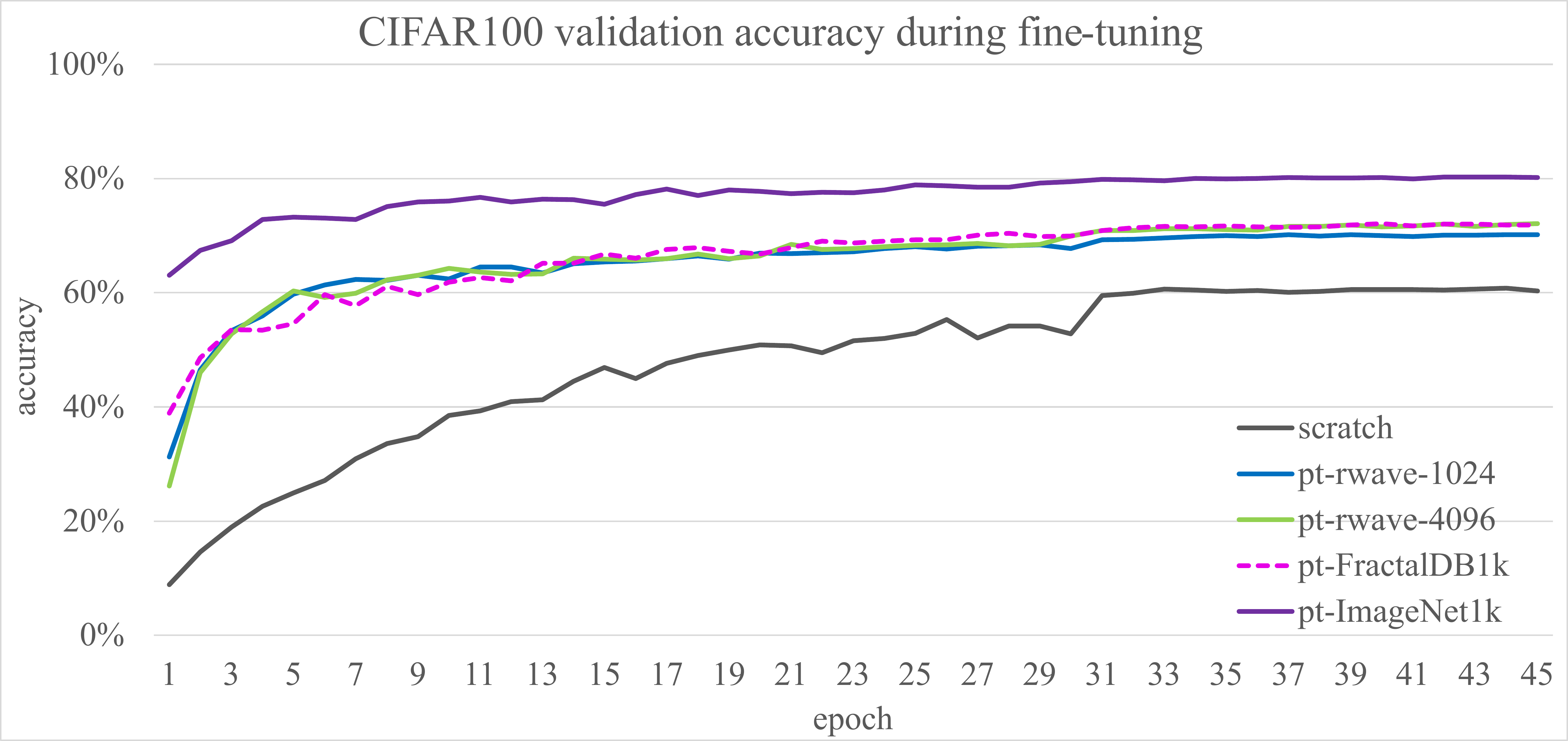}
    \caption{Comparison of ResNet50 training on CIFAR100. scratch: no pre-training was done. pt-: pre-trained on the mentioned dataset. Default parameters of the FractalDB pipeline were used.}
    \label{fig:fig4}
\end{figure}

\begin{table}[H]
    \caption{Accuracies and losses for tested ResNet50 training strategies with an initial learning rate of 0.1 during fine-tuning. For testing, the validation dataset has been used. \textbf{Best} for the last dataset in the training strategy is marked in bold.}
    \label{tab:acc2}
    \centering
    \begin{tabular}{rll}
        \toprule
        \it training datasets & Test accuracy $\uparrow$ & Test Loss $\downarrow$ \\
        \midrule
        \it ImageNet1k & \textbf{72.42}\% &  1.218\\
        \it rwave-4096 $\to$ \it ImageNet1k & 71.77\% & \textbf{1.201} \\
        \it rwave-1024 $\to$ \it ImageNet1k & 71.67\% & 1.247 \\
        \it FractalDB1k $\to$ \it ImageNet1k & 71.68\% & 1.244 \\
        \bottomrule
    \end{tabular}  
\end{table}
\textbf{Generalization Time for CIFAR100:} In comparison to training from scratch, we observe faster generalization when pre-training on retinal waves and fine-tuning on CIFAR100. The generalization process is very similar to FractalDB1k pre-training. ImageNet1k pre-training enables the fastest generalization. A comparison of training and validation accuracy for the first 45 epochs of training is shown in Figure~\ref{fig:fig4}. This again matches the previous observations reported for FractalDB and gives a strong argument why such a pre-training might be a sensible choice – using such pre-training later processing steps can be built more efficiently and the training data seen later does not have to first train the early processing steps. This could partially explain the high degree of data-efficiency we observe in the training of the human visual system.

\textbf{Brain-Score} combines many vision benchmarks for artificial neural networks and computes scores to measure how well the ANNs match physiological measurements of each region of the ventral stream of the primate brain \cite{SchrimpfKubilius2018BrainScore, Schrimpf2020integrative}. To evaluate how well our models match regions in the ventral stream of the primate brain (V1, V2, V4 and IT), we compare the ceiled Brain-Score results of multiple models in Table \ref{tab:brainscore}. The Behavior score is obtained by evaluating the model on various behavioral benchmarks \cite{SchrimpfKubilius2018BrainScore, Schrimpf2020integrative}. 

\begin{table}
    \caption{Brain-Score results for tested ResNet50 training strategies. Scores range from 0 to 1, higher is better. Currently best performing model on BrainScore \cite{riedel2022bag} achieves an average score of .465. \textbf{Highest score} in each category is marked in bold. (For these training strategies, the default configuration of the FractalDB training pipeline was used (initial learning rate of 0.01). Only for training strategies containing the ImageNet1k dataset, the initial learning rate was set to 0.1 during fine-tuning.)}
    \label{tab:brainscore}
    \centering
    \begin{tabular}{rccccccc}
        \toprule
        \it training datasets & Avg$\uparrow$ & V1$\uparrow$ & V2$\uparrow$ & V4$\uparrow$ & IT$\uparrow$ & Behavior$\uparrow$\\
        \midrule
        \it rwave-1024 & .230 & .437 & .217 & .252 & .185 & .059\\
        \it rwave-4096 & .239 & .434 & \textbf{.226} & \textbf{.279} & \textbf{.190} & .065\\
        \it FractalDB1k & \textbf{.267} & \textbf{.572} & .222 & .250 & \textbf{.190} & \textbf{.101} \\
        \midrule
        \it CIFAR100 & .216 & .389 & .154 & .305 & .225 & .010 \\
        \it rwave-1024 $\to$ \it CIFAR100 & \textbf{.290} & .455 & \textbf{.267} & .392 & \textbf{.276} & .060\\
        \it rwave-4096 $\to$ \it CIFAR100 & .282 & .444 & .236 & \textbf{.399} & .266 & .067 \\
        \it FractalDB1k $\to$ \it CIFAR100 & .285 & \textbf{.506} & .222 & .372 & .251 & \textbf{.073} \\
        \midrule
        \it ImageNet1k $\to$ \it CIFAR100 & \textbf{.313} & .458 & \textbf{.277} & \textbf{.437} & \textbf{.296} & \textbf{.097} \\
        \midrule
        \it ImageNet1k & \textbf{.414} & \textbf{.538} & .315 & \textbf{.497} & .378 & .342 \\
        \it rwave-1024 $\to$ \it ImageNet1k & .399 & .524 & .314 & .491 & \textbf{.390} & .277\\
        \it rwave-4096 $\to$ \it ImageNet1k & .411 & .517 & .308 & .488 & .383 & \textbf{.361}\\
        \it FractalDB1k $\to$ \it ImageNet1k & .407 & .525 & \textbf{.324} & .489 & \textbf{.390} & .306 \\
        \bottomrule
    \end{tabular}
\end{table}

Models pre-trained on retinal waves and fine-tuned on CIFAR100 show increased scores in all regions compared to training from scratch. This supports strongly that pre-training ANNs with retinal waves is beneficial and could lead to better models of the primate visual system. Especially the improved representation in early layers makes sense in our eyes, as retinal waves might enable to learn those features in a less noisy and cluttered way.

Interestingly, models fine-tuned on ImageNet1k do not show large differences in the BrainScores, which could be explained by the higher initial learning rate (0.1) which was used during fine-tuning. 

The Brain-Scores have been obtained by utilizing the base model of Brain-Score; Layers of the CNN are matched automatically to regions of the primate ventral stream when using the base model. Between different models, the different layers were matched inconsistently, which could explain some score discrepancies.

\section{Limitations}
The parameters for the retinal wave simulator used for our experiment are based on animal experiments and primate or human retinal waves might behave slightly different. We however do not have as much observations for human retinal waves. Using observations of human retinal waves could yield even more insights in how the human brain behaves regarding vision.

We only investigated pre-training in a fully supervised fashion. One could argue, that unsupervised learning would be more biologically plausible.  By releasing our data generator and training data we also enable other researchers to try different routes and training schemes.

As the visual cortex has not yet fully developed when retinal waves do occur, a fixed ANN architecture is a limiting factor. An interesting extension of our approach could go in the direction of neural architecture search.

\section{Conclusion}
We have trained ANNs based on retinal waves and observe not only the emergence of strong and biologically plausible features in early layers, but also a match to primate brain activity of our pre-trained networks and an induced performance gain similar to that of state-of-the-art pre-training dataset FractalDB \cite{KataokaACCV2020, KataokaIJCV2022}.
ReWaRD enables future research to study human visual development with a curriculum learning approach. The approach could be extended, e.g. by expanding the training strategy with different pre-training steps for each stage of retinal waves or even more steps that mimic different stages of early visual experience, e.g. using the SAYCam dataset \cite{SAYCAM}.
An additional benefit we would like to emphasize is the absence of bias towards faces, animals or similar object categories in our datasets which could be very beneficial for applications in both neuroscience and computer vision, as it might reduce misleading results arising from a biased training diet and might reduce bias in downstream applications after fine-tuning.

Furthermore, the pre-trained networks \cite{cappell2023pretrainedANNs} can be fine-tuned for many different applications, as they are made publicly available. Also, the generated retinal wave datasets \cite{cappell2023rwave-1024, cappell2023rwave-4096} are made publicly available. 

Our code base for generating retinal waves, training the models and for analyzing (BrainScore) and visualizing trained networks along with further example feature visualizations is made publicly available as well. \cite{cappell2023rwavesim, cappell2023ReWaRDproj} 

ReWaRD Project Page: \href{https://github.com/bennyca/reward}{\url{https://github.com/bennyca/reward}}

Retinal Wave Simulator Code \cite{cappell2023rwavesim}: \href{https://zenodo.org/records/8150777}{\url{https://zenodo.org/records/8150777}} \newline
ReWaRD Code  \cite{cappell2023ReWaRDproj}: \href{https://zenodo.org/records/10148723}{\url{https://zenodo.org/records/10148723}}\newline
rwave-1024 dataset \cite{cappell2023rwave-1024}: \href{https://zenodo.org/records/7811860}{\url{https://zenodo.org/records/7811860}}\newline
rwave-4096 dataset \cite{cappell2023rwave-4096}: \href{https://zenodo.org/records/7779499}{\url{https://zenodo.org/records/7779499}}\newline
ANNs pre-trained and fine-tuned \cite{cappell2023pretrainedANNs}: \href{https://zenodo.org/records/10148752}{\url{https://zenodo.org/records/10148752}}\newline \\

\begin{ack}
This project was funded by the Federal Ministry of Education and Research of Germany (BMBF) project 01IS22082 (IRRW) and the FAU Emerging Talent Initiative (ETI). 
The authors gratefully acknowledge support by the Google Cloud Research Credits program award GCP19980904. 
The authors are responsible for the content of this publication.

\end{ack}

{\small
\bibliographystyle{ieee_fullname}
\bibliography{arxiv}
}
\end{document}